\documentclass[conference]{IEEEtran}
\IEEEoverridecommandlockouts
\usepackage{tabularx}
\usepackage{cite}
\usepackage{amsmath,amssymb,amsfonts}
\usepackage{graphicx}
\usepackage{textcomp}
\usepackage{xcolor}
\usepackage{varwidth}
\def\BibTeX{{\rm B\kern-.05em{\sc i\kern-.025em b}\kern-.08em
    T\kern-.1667em\lower.7ex\hbox{E}\kern-.125emX}}
\begin{document}

\title{Semi-Supervised Learning-Based Genetic Biomarkers Dataset for Multiple-Stage Hepatocellular Carcinoma Prediction}

\author{
\IEEEauthorblockN{
Ahmed Ammar Kubba\IEEEauthorrefmark{1},
Manar Abu Talib\IEEEauthorrefmark{1},
Jibran Sualeh Muhammad\IEEEauthorrefmark{2},
Ali Bou Nassif\IEEEauthorrefmark{3}\\
Abdalla Sayed Mohamed\IEEEauthorrefmark{1},
Darko Castven\IEEEauthorrefmark{4},
Jens U. Marquardt\IEEEauthorrefmark{4}
}

\IEEEauthorblockA{\IEEEauthorrefmark{1}
Department of Computer Science, University of Sharjah, Sharjah, UAE\\
Emails: \{U23103280, mtalib, U22103623\}@sharjah.ac.ae
}

\IEEEauthorblockA{\IEEEauthorrefmark{2}
Department of Biomedical Sciences, University of Birmingham, Birmingham, United Kingdom\\
Email: dr.jibran@live.com
}

\IEEEauthorblockA{\IEEEauthorrefmark{3}
Department of Computer Engineering, University of Sharjah, Sharjah, UAE\\
Email: anassif@sharjah.ac.ae
}

\IEEEauthorblockA{\IEEEauthorrefmark{4}
Department of Medicine I, University Medical Center Schleswig-Holstein,
Campus L\"ubeck, L\"ubeck, Germany\\
Emails: \{Darko.Castven, Jens.Marquardt\}@uksh.de
}
}

\maketitle

\begin{abstract}
Liver cancer is a complex disease responsible for a high number of deaths across the globe each year, making automated solutions for liver cancer classification urgent. The most common form of liver cancer is hepatocellular carcinoma (HCC), accounting for over 90\% of liver cancer cases. There is a distinct lack of publicly available HCC datasets utilizing genomic data, which is necessary for training artificial intelligence (AI) models for automated HCC classification. This study proposes constructing a multi-stage HCC dataset using XGBoost and Semi-Supervised learning on three separate datasets of genomic biomarkers, utilizing their existing labels in the Semi-Supervised learning process to label the proposed dataset. The proposed dataset consists of 770 patient samples in total, categorized into five classes that represent normal tissue alongside different stages of HCC. Each sample in the dataset consists of 11,150 different gene expression levels. The XGBoost model demonstrated a final classification accuracy of 96.5\% during the Semi-Supervised learning process.
\end{abstract}
\renewcommand{\IEEEkeywordsname}{Keywords}

\section{Introduction and Related Work}

Liver cancer was ranked sixth in incidence with 905,677 cases and fourth in mortality with 830,180 deaths globally in 2023. With an expected incidence surpassing 1 million cases by 2025, it remains a major global health concern \cite{masuzaki2023liver}. Hepatocellular Carcinoma (HCC) is a type of cancer that develops from mutations of liver cells called hepatocytes. HCC is responsible for over 90\% of liver cancer cases \cite{farasati2023unresectable}. Modern research has focused extensively on non-invasive biomarkers and imaging techniques like MRI. Machine learning models for diagnosing liver cancer have also been of interest to researchers. These innovations have significantly advanced HCC research by aiding in prediction of tumour recurrence and identification \cite{addissouky2024latest}. Recent studies have demonstrated correlations between HCC and various biomarkers for identification and progression. For instance, anuradha et al. \cite{budhu2013integrated} identified 28 metabolites and 169 genes that correlate with aggressive HCC progression and outcomes, with supporting data made publicly available. During cancer development, liver cells often present distinct molecular signatures, and release certain tumour-associated molecules into body fluid, e.g., blood, urine or stool, that could be monitored for the onset or progression of HCC. There has been a rise in technologies that operate using techniques like chemiluminescence immunoassay, enzyme-linked immunosorbent assay, immunosensor, proteomics, and liquid biopsy \cite{pan2020biomarkers}. Different multi-omics approaches have also been explored in the literature \cite{chen2022potential}.

In the early stages of HCC, tumor sizes are typically less than 5 cm in diameter and range from one to three tumors. Early HCC rarely shows significant symptoms or liver dysfunction. There is also no significant spread to nearby blood vessels or organs, resulting in a better prognosis. Early detection is, therefore, crucial for patient survival. In contrast, advanced HCC involves larger and multiple tumors, often with invasion into blood vessels or other organs, and has a poor prognosis \cite{kuo2021factors}. Low-grade dysplastic nodules (LGDN) are early precancerous lesions that resemble regenerative nodules with mild atypia, indicating slight abnormalities in cell structure. They also show less aggressive behaviour and fewer molecular changes associated with cancer progression. High-grade dysplastic nodules (HGDN), on the other hand, exhibit more pronounced atypia and are considered more advanced precancerous lesions with a higher risk of progression to HCC. HGDN often display molecular alternations which is common in early stages of HCC, making them critical targets for early cancer detection. Various biomarkers that can help in distinguishing between LGDN and HGDN \cite{duan2020differentiation}.

Das et al. \cite{das2022deep} introduced a novel approach for detecting cancer genes using different DNA sequences. The authors used the VGG16 deep learning model and the NCBI dataset. Their training data consisted of sequences of four healthy genes and four HCC genes, using three numerical mapping techniques to digitize the gene sequences. The genes were examined in both one-dimensional and two-dimensional forms using convolutional neural networks (CNN). Their CNN model achieved an 80.36\% accuracy in the one-dimensional form. The SVM and VGG16 models achieved a high 98.86\%, and 100\% accuracy with fine-tuned VGG16 layers. This method effectively extracted features to distinguish HCC from normal liver gene sequences, allowing for broader applications with larger datasets and different cancer types. One critical limitation to this approach was the absence of enough genes and data in the used dataset, which was not large enough to provide a sufficient training set to design a new CNN model. Other studies in the literature have also noted a general limitation in available liver cancer medical data and small sample sizes, leading to inevitable constraints on liver cancer research \cite{chen2024icycle,yang2024performance,zhi2025parameter,oh2024identification}.

In this study, we construct a dataset containing multi-omics genetic data for multi-stage HCC classification for five different developmental hepatocellular carcinoma stages for the patient. The main contributions of this work are as follows: This study utilizes semi-supervised learning based on several different multi-omics datasets to construct the primary HCC dataset used to train our deep learning model, which addresses the challenge of limited and small HCC datasets which other studies tend to face. Additionally, our HCC dataset contains five different categories for patient tissue samples representing different stages of HCC, with 770 samples each containing 11,150 genomic expressions, making it more detailed compared to other publicly available datasets.

This paper is structured in the following manner: The first section introduces the research subject, HCC data and stages of development, and relevant background information, in addition to a review of the related literature, highlighting key drawbacks and limitations in the existing studies. The following section covers the Methodology, including the datasets, pre-processing and the semi-supervised learning. The third section covers the results and their discussion, and the fourth section contains the conclusion and future work of the project.

\section{Methodology}
\begin{figure}
\includegraphics[width=0.5\textwidth]{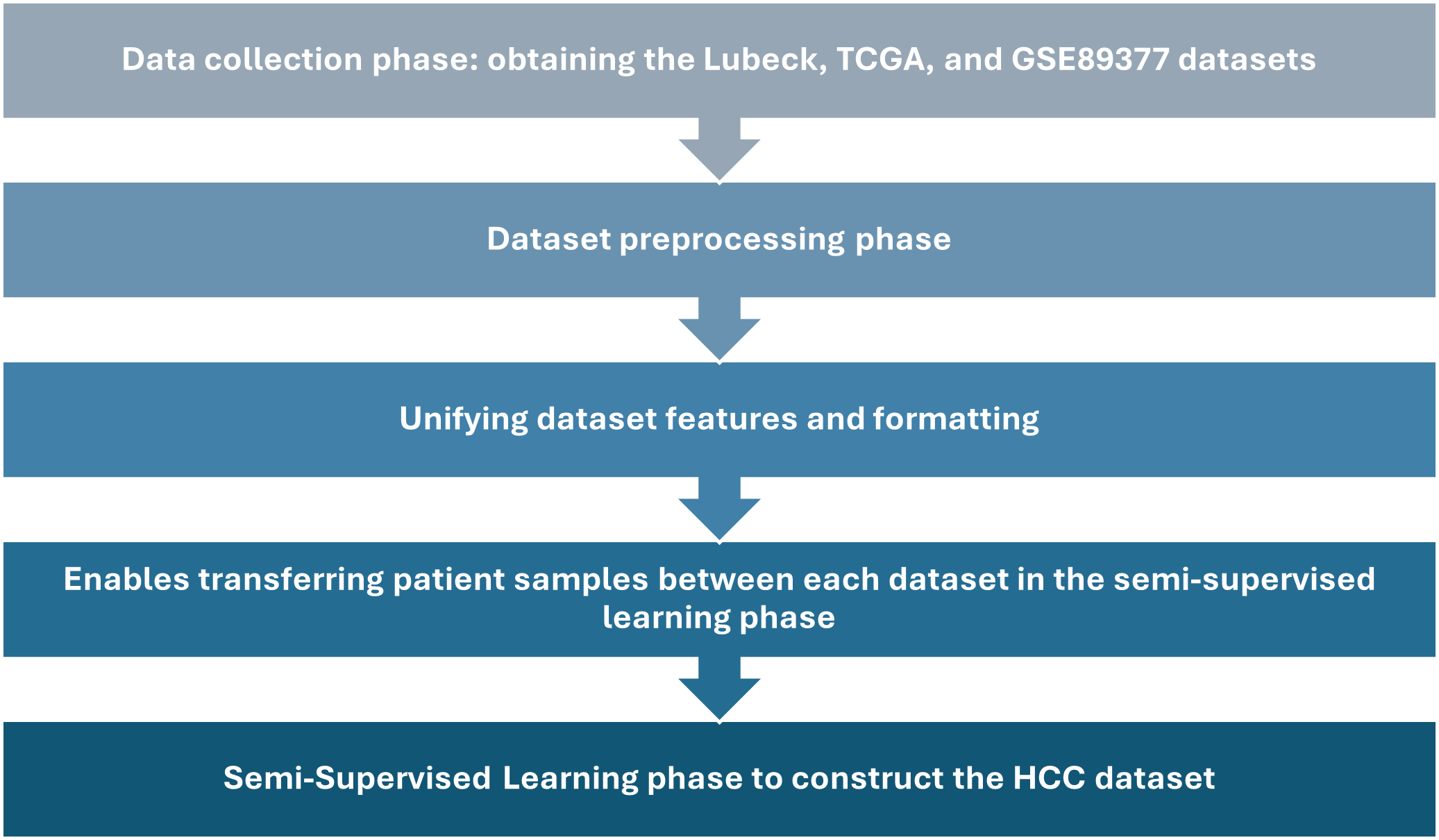}
\caption{Visual summary of the paper methodology, outlining each step in the process of constructing the HCC dataset.} \label{fig1}
\end{figure}

This section describes the methodology of the project in detail, which consists of obtaining the base dataset from medical domain experts in addition to dataset pre-processing and semi-supervised learning using three different datasets to build the final HCC dataset. Figure~\ref{fig1} illustrates the overall steps of the methodology, consisting of the initial HCC data collection phase from the three datasets, after which the data is pre-processed and properly formatted. Finally, the semi-supervised learning phase is conducted to construct the HCC dataset. The next sub-sections will elaborate on each phase in Fig.~\ref{fig1}.
\subsection{Lubeck Dataset and Target Classes}
The five classes or ground-truth labels in the Lubeck dataset are described in Table~\ref{classes1}, which defines the patient tissue samples as five distinct categories: "Surrounding Liver (sl)," which indicates no HCC gene expressions; "Early hepatocellular carcinoma (ehcc)," which represents early-stage HCC diagnosis; "Progressed hepatocellular carcinoma (phcc)," which represents progressed HCC diagnosis; "Low-grade dysplastic nodules (lgdn)," which is associated with a lower risk of developing HCC \cite{wu2024feeding}; and "High-grade dysplastic nodules (hgdn)," which is associated with a higher risk of developing HCC. This private dataset serves as the basis for our HCC dataset, as it contains the most detailed labels for HCC stages in comparison to the two other publicly available datasets and thus represents the starting point for the semi-supervised learning phase.
\begin{table}[ht]
\centering
\caption{Description of the dataset classes, consisting of SL, EHCC, PHCC, LGDN, and HGDN.}
\label{classes1}
\begin{tabularx}{\columnwidth}{|p{2.2cm}|X|}
\hline
\textbf{Class} & \textbf{Description} \\
\hline
Surrounding Liver (SL) & This class represents patient tissue samples taken from the surrounding liver area of the patient, which contains gene expressions that indicate no HCC in the patient. \\
\hline
Early Hepatocellular Carcinoma (EHCC) & This class represents samples of patients with a diagnosis of early-stage HCC. \\
\hline
Progressed Hepatocellular Carcinoma (PHCC) & This class is assigned to patients with a diagnosis of progressed HCC. \\
\hline
Low-Grade Dysplastic Nodules (LGDN) & Dysplastic nodules are associated with a higher risk of developing HCC; low-grade dysplastic nodules represent a much lower risk.\\
\hline
High-Grade Dysplastic Nodules (HGDN) & High-grade dysplastic nodules indicate a high risk of developing HCC, as represented by this class. \\
\hline
\end{tabularx}
\end{table}

\subsection{Dataset Pre-Processing}
The main dataset used in training the artificial intelligence (AI) model was built using a semi-supervised learning approach based on three source datasets that are detailed in this section.

\textbf{1. Lubeck Dataset.}
This private dataset contains the original patients' sample data collected by a team of medical domain experts in the Lubeck university lab. It consists of 29 samples in total, with 16,381 gene expression features for each patient. It also consists of the five class labels for categorizing each of the 29 patients: sl, ehcc, phcc, lgdn, and hgdn.

\textbf{2. The Cancer Genome Atlas (TCGA) Dataset.}
Launched in 2006 as a collaborative endeavour between the National Cancer Institute (NCI) and the National Human Genome Research Institute, The Cancer Genome Atlas \cite{baird2024gs} is a cancer genomics initiative which has profiled over 20,000 primary cancer and corresponding normal samples across 33 different cancer types. This is a public HCC dataset which consists of 782 patient samples and 16,382 gene expression features for each patient. The dataset is categorized into three main class labels: Normal, HCC, and Transition.

\textbf{3. GSE89377 Dataset.}
GSE89377 is a gene expression dataset (publicly available in the National Center for Biotechnology Information Gene Expression Omnibus) which contains 107 patient liver tissue samples that cover nine stages of HCC development, with normal liver tissue used as control and around 48,000 gene expression features for each sample \cite{shen2018barrier}.

The GSE89377 dataset required several preprocessing steps to be utilized in this work. Firstly, it was necessary to match the gene keywords to their corresponding symbols in the Illumina HumanHT-12 V3.0 Expression BeadChip (GPL6947), which is a widely-used microarray platform used for gene expression profiling in human samples \cite{dong2023genetic}. This was done to ensure computability before adding samples from the GSE89377 dataset to the Lubeck and TCGA datasets. The nine class labels covered in the dataset for patient classification are as follows: `Normal', `Chronic Hepatitis with Low/High Grade', `Cirrhosis', `Dysplastic Nodules with Low/High Grade', `Early HCC', `HCC TG1/TG2/TG3'. Additionally, only the gene expression features in common between all three datasets were kept, and all other features were discarded. This was done to enable combining the samples from all dataset sources into one HCC dataset.

\subsection{Semi-Supervised Learning}
The proposed dataset was created using a semi-supervised machine learning approach. Semi-supervised learning enhances the performance of traditional supervised learning, which requires labeled data samples for model training \cite{soudan2025scalability}, by enabling researchers to make use of unlabelled data. In recent years, it has attracted increasing interest from researchers as one possible approach to reducing dependence on labelled datasets \cite{han2024deep}. As the GSE89377 and TCGA datasets contain different labels than the five target labels in the Lubeck dataset, the usage of semi-supervised learning helped us overcome this obstacle.

The appropriate samples from the GSE89377 dataset were first added to the Lubeck and TCGA datasets according to their class labels, which were mapped to the five class labels in the other datasets based on medical domain expert analysis and insight. The liver cirrhosis samples were reclassified as `Normal' samples and added to the TCGA dataset, whereas the chronic hepatitis samples were discarded as they are unrelated to hepatocellular carcinoma. The `Normal' samples were classified as `sl' and added to the Lubeck dataset alongside the dysplastic nodules samples as lgdn/hgdn and Early HCC samples as `ehcc'. The HCC Tumor Grade 1 (TG1) samples were classified as HCC and added to the TCGA dataset, whereas the TG2 and TG3 samples were mapped to the Lubeck dataset as progressed HCC (phcc) samples.
\begin{figure}
\includegraphics[width=0.5\textwidth]{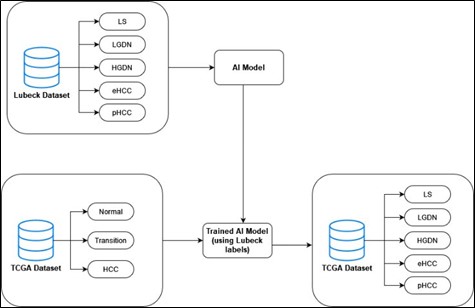}
\caption{Illustration of the Semi-Supervised learning process, using three different sources for constructing the HCC dataset.} \label{fig2}
\end{figure}
The semi-supervised learning approach, as visualized in Figure~\ref{fig2} and summarized in Figure~\ref{fig3}, consisted first in training an XGBoost model on the base Lubeck dataset, which contains the five target classes. Afterwards, the trained model predicted the classes in the TCGA dataset. High-confidence predictions (as in, predictions with a high probability score) which match the real TCGA class labels were added to the Lubeck dataset as new samples. Finally, the new model was trained on the expanded Lubeck dataset to compare its new performance to the original dataset, with a stratified training and testing dataset split of 70\% and 30\%, respectively.

These steps were repeated until convergence. The convergence criteria was defined by the expanded dataset leading to the model's performance degrading or not changing with further iterations. The following are the basic conditions for adding a prediction made on the TCGA dataset to the Lubeck dataset, made with the advice of medical domain experts: 1) It must satisfy the defined prediction confidence threshold (40\%); 2) If the true label is ``HCC'', the prediction must be either ``ehcc'' or ``phcc''; 3) If the true label is ``Transition'', the prediction must be either ``lgdn'' or ``hgdn''; 4) If the true label is ``Normal'', the prediction must be ``sl''.
\begin{figure}
\includegraphics[width=0.5\textwidth]{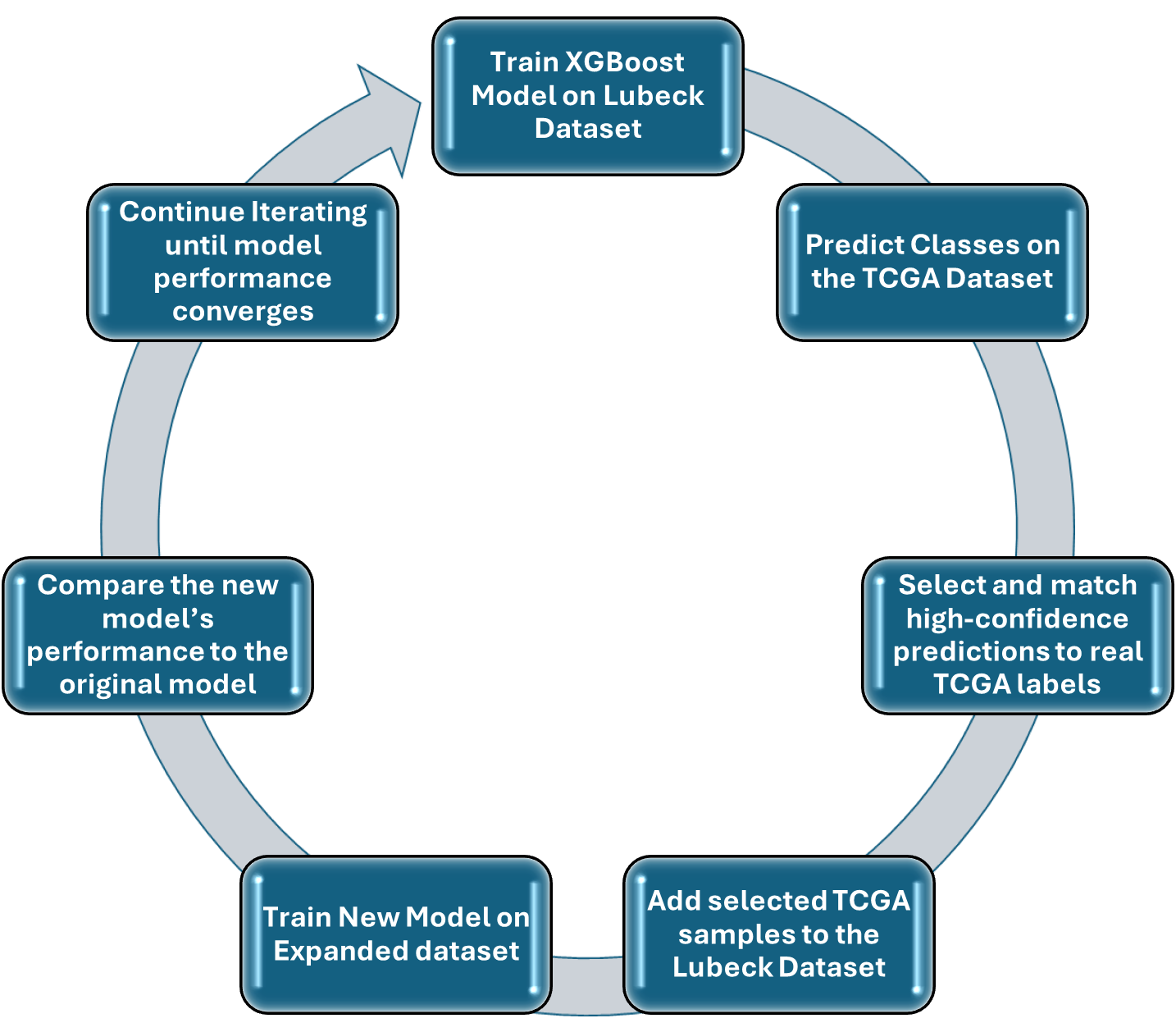}
\caption{Visual summary of the Semi-Supervised Training process, using the XGBoost model to create the HCC dataset.} \label{fig3}
\end{figure}

\section{Results and Discussion}
This section discusses the results of the semi-supervised learning process, which consist of the final XGBoost model and its performance, in addition to the HCC dataset and its class distribution and genomic features.
\subsection{XGBoost Model}
\begin{figure}
\includegraphics[width=0.5\textwidth]{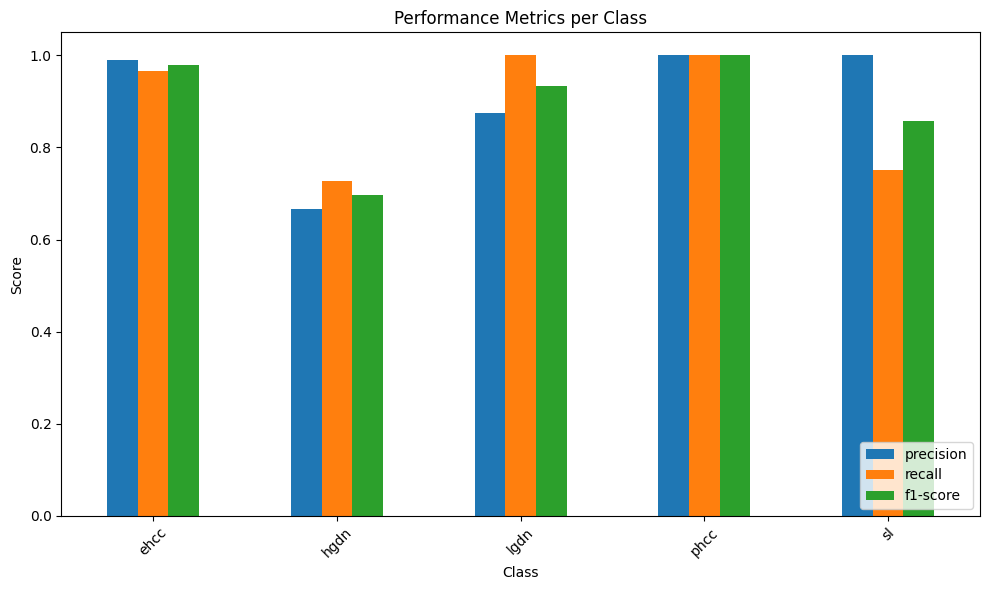}
\caption{Visual comparison of the XGBoost model's performance across the HCC dataset's five classes.} \label{permet}
\end{figure}

\begin{figure}
\includegraphics[width=0.5\textwidth]{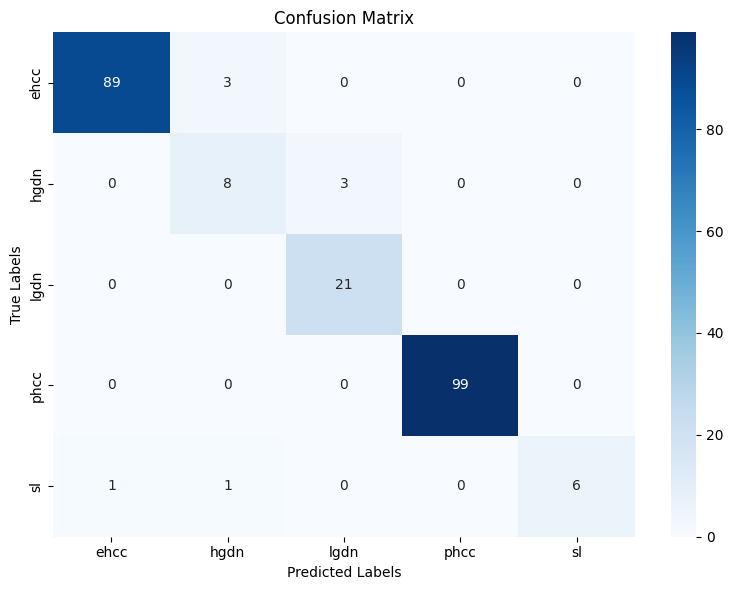}
\caption{Confusion matrix for the XGBoost model, visualizing the misclassifications of the model across the five classes.} \label{confmat}
\end{figure}

The final performance metrics of the XGBoost model that was trained on the HCC dataset using all the gene features were recorded in Table~\ref{classes}. The XGBoost model demonstrates strong performance, achieving \textbf{96.5\% accuracy}, \textbf{90.6\% precision}, \textbf{88.9\% recall}, and an \textbf{F1 score of 89.3\%}, showing a balanced ability to correctly identify multiple stages of HCC samples while minimizing false positives. The low cross-entropy loss of \textbf{0.1015} indicates that the model is highly confident in its predictions. However, the dataset class imbalance, noted in the Conclusion section, must also be taken into consideration when comparing the performance of the XGBoost model with other models, as it was mainly used for constructing the dataset during the semi-supervised learning process. The minor trade-off between precision and recall means that the XGBoost model leans towards reducing false positives, which can be desirable in medical applications. The model's performance across the five different classes can also be observed in Figure~\ref{permet}, and the confusion matrix in Figure~\ref{confmat}.
\begin{table}[ht]
\centering
\caption{This table records the XGBoost performance metrics, consisting of accuracy, precision, recall, and F1 score on the HCC dataset.}
\label{classes}
\begin{tabular}{|l||l|l|l|l|}
\hline
   Class&Precision& Recall&F1-Score& Support\\
\hline
    ehcc&98.9\%& 96.7\%&97.8\%& 92\\
\hline
    hgdn&66.7\%& 72.7\%&69.6\%& 11\\
\hline
    lgdn&87.5\%& 100\%&93.3\%& 21\\
\hline
    phcc&100\%& 100\%&100\%& 99\\
\hline
    sl&100\%& 75\%&85.7\%& 8\\
\hline
 Accuracy& 96.5\%& 96.5\%& 96.5\%&96.5\%\\\hline
 Macro Avg& 90.6\%& 88.9\%& 89.3\%&231\\\hline
 Weighted Avg& 96.8\%& 96.5\%& 96.6\%&231\\\hline
\end{tabular}
\end{table}
\subsection{HCC Dataset}
\begin{figure}
\includegraphics[width=0.5\textwidth]{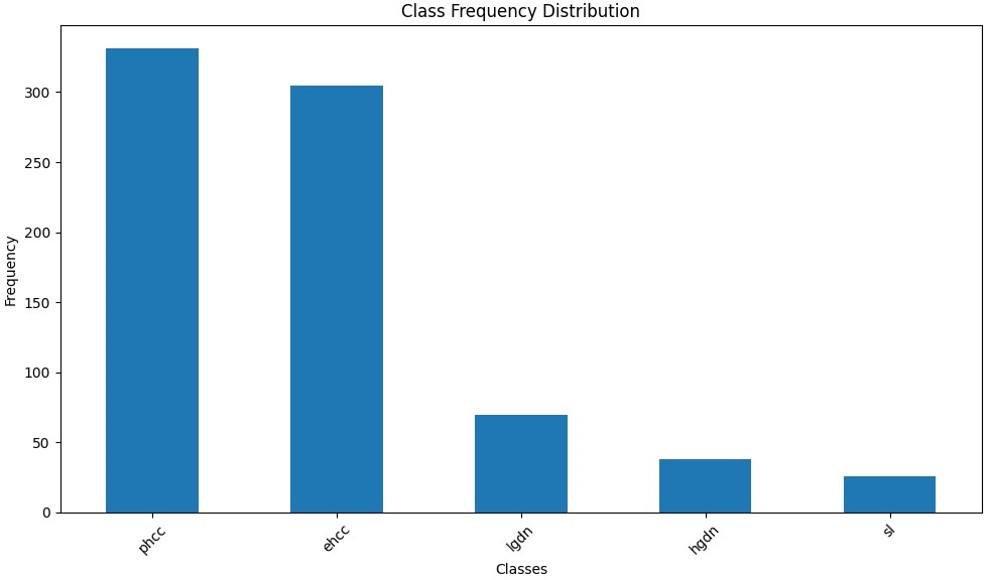}
\caption{This graph illustrates the frequency distribution of the different classes in the HCC dataset.} \label{fig4}
\end{figure}

\begin{figure}
\includegraphics[width=0.5\textwidth]{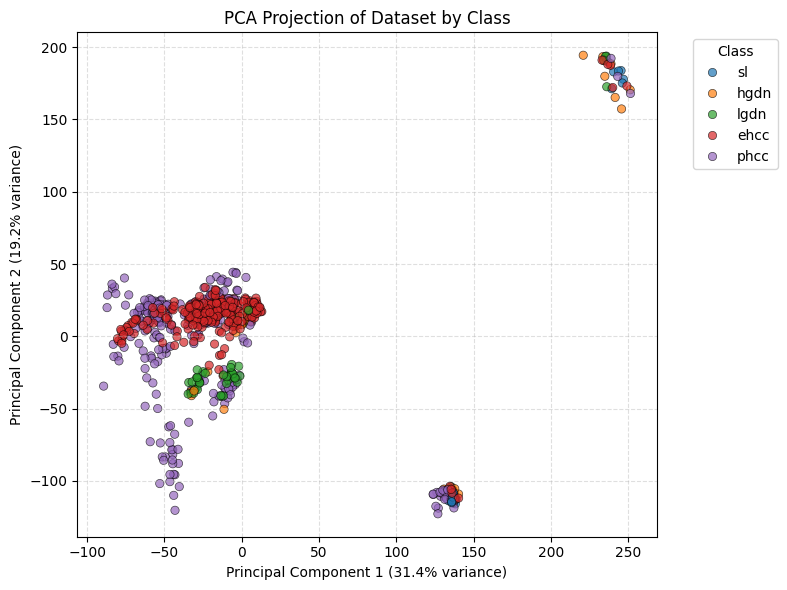}
\caption{PCA plot of the HCC dataset, illustrating the distribution of the classes in feature space.} \label{pca}
\end{figure}

The constructed HCC dataset consists of 770 total samples taken from different patients representing their multi-omics genetic expression data. The frequency distribution of the dataset's classes can be observed in Figure~\ref{fig4}, which shows a significant imbalance, where the phcc and ehcc classes have the highest number of samples while lgdn, hgdn, and sl are underrepresented in comparison. This imbalance is a common issue in medical datasets beyond just liver cancer data \cite{alsalama2024classification}, and can impact model performance if not taken into consideration, potentially leading to model bias toward the majority classes during prediction. Each patient has 11,150 feature columns that indicate the different genomic expression levels of various genes, which can be potentially used to make an informed prediction of the HCC diagnosis stage of the patient. It can be observed in Figure~\ref{fig4} that the most frequent samples in the dataset belong to the `phcc' and `ehcc' classes, whereas the least frequent samples belong to the `sl' class with less than 50 samples belonging to that class. A principal component analysis (PCA) plot of the dataset can be observed in Figure~\ref{pca}, which visualizes how the five classes are distributed in feature space.
\section{Conclusion}
This work presents a comprehensive dataset for hepatocellular carcinoma analysis by constructing a relatively large collection of high-dimensionality data using a semi-supervised learning-based approach based on one private dataset in addition to two public datasets, mapping the labels in the latter two datasets to the labels of the Lubeck dataset. The constructed HCC dataset can support advancements in medical research, early diagnosis, and treatment planning by enabling AI-driven models to enhance accuracy and efficiency in liver cancer detection. Researchers can use this data to develop automated classification systems and predictive models that can contribute to improved liver cancer patient outcomes.

Limitations and future work include the dataset imbalance limitation, which can be addressed in future research through techniques like class weighting or synthetic minority over-sampling to help mitigate the effects of class imbalance and improve model generalization. Additionally, further validation using external datasets and probability calibration checks can be conducted to further ensure model robustness and generalizability. Further interdisciplinary collaboration between medical professionals and machine learning experts can lead to improving the dataset's quality and sample diversity, in addition to finding relevant features or genes in the dataset which can be clinically validated as novel contributions to medical research.

\section*{Declaration of Interests}

The authors declare that they have no known competing financial interests or personal relationships that could have appeared to influence the work reported in this paper.

\bibliographystyle{IEEEtran}
\bibliography{ref}  

@misc{masuzaki2023liver,
  title={Liver cancer: Improving standard diagnosis and therapy},
  author={Masuzaki, Ryota},
  journal={Cancers},
  volume={15},
  number={18},
  pages={4602},
  year={2023},
  publisher={MDPI}
}

@article{farasati2023unresectable,
  title={Unresectable hepatocellular carcinoma: a review of new advances with focus on targeted therapy and immunotherapy},
  author={Farasati Far, Bahareh and Rabie, Dorsa and Hemati, Parisa and Fooladpanjeh, Parastoo and Faal Hamedanchi, Neda and Broomand Lomer, Nima and Karimi Rouzbahani, Arian and Naimi-Jamal, Mohammad Reza},
  journal={Livers},
  volume={3},
  number={1},
  pages={121--160},
  year={2023},
  publisher={MDPI}
}

@article{addissouky2024latest,
  title={Latest advances in hepatocellular carcinoma management and prevention through advanced technologies},
  author={Addissouky, Tamer A and Sayed, Ibrahim El Tantawy El and Ali, Majeed MA and Wang, Yuliang and Baz, Ayman El and Khalil, Ahmed A and Elarabany, Naglaa},
  journal={Egyptian Liver Journal},
  volume={14},
  number={1},
  pages={2},
  year={2024},
  publisher={Springer}
}

@article{pan2020biomarkers,
  title={Biomarkers in hepatocellular carcinoma: current status and future perspectives},
  author={Pan, Yasi and Chen, Huarong and Yu, Jun},
  journal={Biomedicines},
  volume={8},
  number={12},
  pages={576},
  year={2020},
  publisher={MDPI}
}

@article{chen2022potential,
  title={Potential biomarkers for liver cancer diagnosis based on multi-omics strategy},
  author={Chen, Fanghua and Wang, Junming and Wu, Yingcheng and Gao, Qiang and Zhang, Shu},
  journal={Frontiers in Oncology},
  volume={12},
  pages={822449},
  year={2022},
  publisher={Frontiers Media SA}
}

@article{kuo2021factors,
  title={Factors predicting long-term outcomes of early-stage hepatocellular carcinoma after primary curative treatment: the role of surgical or nonsurgical methods},
  author={Kuo, Ming-Jeng and Mo, Lein-Ray and Chen, Chi-Ling},
  journal={BMC cancer},
  volume={21},
  number={1},
  pages={250},
  year={2021},
  publisher={Springer}
}

@article{duan2020differentiation,
  title={Differentiation of regenerative nodule, dysplastic nodule, and small hepatocellular carcinoma in cirrhotic patients: A contrast-enhanced ultrasound--based multivariable model analysis},
  author={Duan, Yu and Xie, Xiaoyan and Li, Qian and Mercaldo, Nathaniel and Samir, Anthony E and Kuang, Ming and Lin, Manxia},
  journal={European Radiology},
  volume={30},
  number={9},
  pages={4741--4751},
  year={2020},
  publisher={Springer}
}

@article{das2022deep,
  title={Deep transfer learning for automated liver cancer gene recognition using spectrogram images of digitized DNA sequences},
  author={Das, Bihter and Toraman, Suat},
  journal={Biomedical Signal Processing and Control},
  volume={72},
  pages={103317},
  year={2022},
  publisher={Elsevier}
}

@article{chen2024icycle,
  title={ICycle-GAN: Improved cycle generative adversarial networks for liver medical image generation},
  author={Chen, Ying and Lin, Hongping and Zhang, Wei and Chen, Wang and Zhou, Zonglai and Heidari, Ali Asghar and Chen, Huiling and Xu, Guohui},
  journal={Biomedical Signal Processing and Control},
  volume={92},
  pages={106100},
  year={2024},
  publisher={Elsevier}
}

@article{yang2024performance,
  title={Performance analysis of data resampling on class imbalance and classification techniques on multi-omics data for cancer classification},
  author={Yang, Yuting and Mirzaei, Golrokh},
  journal={PLoS One},
  volume={19},
  number={2},
  pages={e0293607},
  year={2024},
  publisher={Public Library of Science San Francisco, CA USA}
}

@article{zhi2025parameter,
  title={Parameter changes and influencing factors in sixty patients with interventional surgery for liver cancer diagnoses},
  author={Zhi, Lin and Chen, Zhi-Hai and Deng, Jun},
  journal={World Journal of Gastrointestinal Surgery},
  volume={17},
  number={2},
  pages={99581},
  year={2025}
}

@article{oh2024identification,
  title={Identification of signature gene set as highly accurate determination of metabolic dysfunction-associated steatotic liver disease progression},
  author={Oh, Sumin and Baek, Yang-Hyun and Jung, Sungju and Yoon, Sumin and Kang, Byeonggeun and Han, Su-hyang and Park, Gaeul and Ko, Je Yeong and Han, Sang-Young and Jeong, Jin-Sook and others},
  journal={Clinical and Molecular Hepatology},
  volume={30},
  number={2},
  pages={247},
  year={2024}
}

@article{wu2024feeding,
  title={Feeding artery: a valuable feature for differentiation of regenerative nodule, dysplastic nodules and small hepatocellular carcinoma in CEUS LI-RADS},
  author={Wu, Jiapeng and Zhao, Qinxian and Wang, Yuling and Xiao, Fan and Cai, Wenjia and Liu, Sisi and Du, Zhicheng and Yu, Xiaoling and Liu, Fangyi and Yu, Jie and others},
  journal={European Radiology},
  volume={34},
  number={2},
  pages={745--754},
  year={2024},
  publisher={Springer}
}

@article{baird2024gs,
  title={Gs-tcga: gene set-based analysis of the cancer genome atlas},
  author={Baird, Tarrion and Roychoudhuri, Rahul},
  journal={Journal of Computational Biology},
  volume={31},
  number={3},
  pages={229--240},
  year={2024},
  publisher={Mary Ann Liebert, Inc., publishers 140 Huguenot Street, 3rd Floor New~…}
}

@article{shen2018barrier,
  title={Barrier to autointegration factor 1, procollagen-lysine, 2-oxoglutarate 5-dioxygenase 3, and splicing factor 3b subunit 4 as early-stage cancer decision markers and drivers of hepatocellular carcinoma},
  author={Shen, Qingyu and Eun, Jung Woo and Lee, Kyungbun and Kim, Hyung Seok and Yang, Hee Doo and Kim, Sang Yean and Lee, Eun Kyung and Kim, Taemook and Kang, Keunsoo and Kim, Seongchan and others},
  journal={Hepatology},
  volume={67},
  number={4},
  pages={1360--1377},
  year={2018},
  publisher={Wiley Online Library}
}

@article{dong2023genetic,
  title={Genetic association between ankylosing spondylitis and major depressive disorders: Shared pathways, protein networks and the key gene},
  author={Dong, Tiantian and Lu, Shiyou and Li, Xuhao and Yang, Jiguo and Liu, Yuanxiang},
  journal={Medicine},
  volume={102},
  number={24},
  pages={e33985},
  year={2023},
  publisher={LWW}
}

@article{han2024deep,
  title={Deep semi-supervised learning for medical image segmentation: A review},
  author={Han, Kai and Sheng, Victor S and Song, Yuqing and Liu, Yi and Qiu, Chengjian and Ma, Siqi and Liu, Zhe},
  journal={Expert Systems with Applications},
  volume={245},
  pages={123052},
  year={2024},
  publisher={Elsevier}
}

@inproceedings{alsalama2024classification,
  title={Classification Based on Association Rules Algorithm for Breast Cancer},
  author={Alsalama, Ali and Kubba, Ahmed and Jamjoum, Ghaith and Al Aghbari, Zaher},
  booktitle={2024 Advances in Science and Engineering Technology International Conferences (ASET)},
  pages={1--6},
  year={2024},
  organization={IEEE}
}

@article{soudan2025scalability,
  title={Scalability and performance evaluation of federated learning frameworks: a comparative analysis},
  author={Soudan, Bassel and Abbas, Sohail and Kubba, Ahmed and Abu Waraga, Omnia and Abu Talib, Manar and Nasir, Qassim},
  journal={International Journal of Machine Learning and Cybernetics},
  volume={16},
  number={5},
  pages={3329--3343},
  year={2025},
  publisher={Springer}
}

@article{budhu2013integrated,
  title={Integrated metabolite and gene expression profiles identify lipid biomarkers associated with progression of hepatocellular carcinoma and patient outcomes},
  author={Budhu, Anuradha and Roessler, Stephanie and Zhao, Xuelian and Yu, Zhipeng and Forgues, Marshonna and Ji, Junfang and Karoly, Edward and Qin, Lun--Xiu and Ye, Qing--Hai and Jia, Hu--Liang and others},
  journal={Gastroenterology},
  volume={144},
  number={5},
  pages={1066--1075},
  year={2013},
  publisher={Elsevier}
}

\end{document}